\documentclass{article}

\usepackage[dblblindworkshop, final]{neurips_2026}
\workshoptitle{ML4Molecules: Agentic Systems for Molecular Sciences}

\usepackage[utf8]{inputenc} 
\usepackage[T1]{fontenc}    
\usepackage{url}            
\usepackage{booktabs}       
\usepackage{amsfonts}       
\usepackage{nicefrac}       
\usepackage{microtype}      
\usepackage{xcolor}
\usepackage[
    colorlinks=true,
    linkcolor=red,
    citecolor=blue,
    urlcolor=blue
]{hyperref}

\usepackage{graphicx} 
\usepackage{subcaption} 
\usepackage{wrapfig} 
\usepackage{amsmath}
\usepackage{afterpage}
\usepackage{multirow}
\usepackage{arydshln}
\usepackage{makecell}
\usepackage{amsthm}

\title{Curriculum Learning for Generative Flexible Docking}

\author{%
  Wenyin Zhou$^{1}\thanks{Corresponding Author}$ \quad Hossein Azizpour$^{1,2}$ \\[1.5ex]
  $^{1}$KTH Royal Institute of Technology \quad
  $^{2}$Science for Life Laboratory  \\
  \texttt{\{wenyinz, azizpour\}@kth.se}
}

\begin{document}

\maketitle

\begin{abstract}
Flexible docking is a highly practical task, aiming to predict the joint configuration of a protein–ligand complex from a protein apo structure and a ligand SMILES string. Diffusion and flow-based generative models have made this task tractable by modeling the posterior distribution over the ligand pose and the protein conformation conditioned on the apo input. The central challenge, however, remains the long-tailed apo-to-holo structure shift, where a small fraction of protein undergo large conformational changes. In this work, motivated by the practical observation that downweighting large shift examples improves generalization, and by the mixed nature of the long tail consisting of genuinely difficult examples and noisy ones, we propose to use curriculum learning for this task, gradually incorporating large conformational change examples during training. With a linear pacing function, we show that an existing method can be strongly enhanced. On the PDBBind ESMFold apo benchmark, with relaxation and confidence-based pose selection, the resulting model GCLDock achieves a $46.52\%$ top-1 success rate (ligand RMSD $<$ 2\AA) and $40.26\%$ when additionally requiring PoseBusters validity, outperforming existing baselines.
\end{abstract}

\section{Introduction}
\begin{wrapfigure}{r}{0.52\textwidth}
    \centering
    \includegraphics[width=\linewidth]{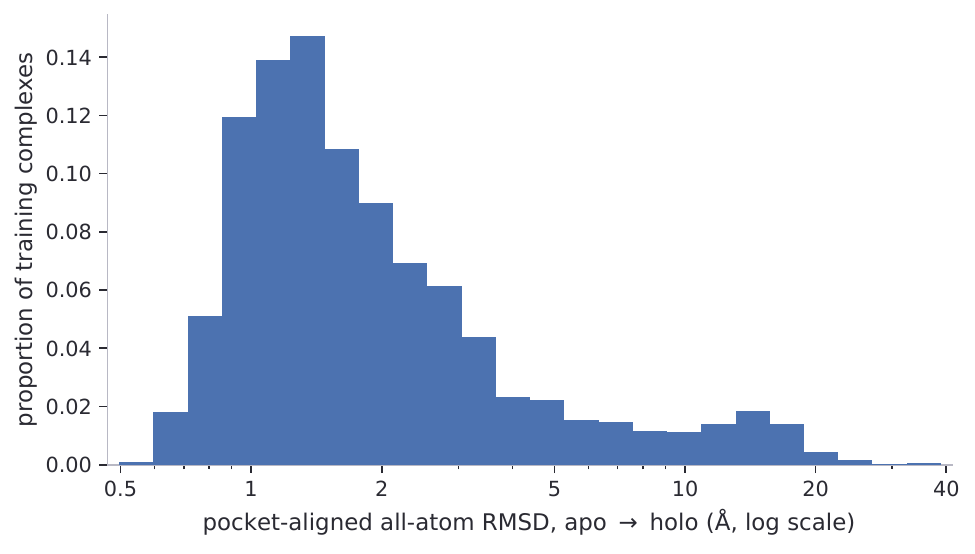}
    \caption{Structure shift of pocket-aligned all-atom RMSD (log scale) between apo and holo conformations on the PDBBind v2020 training set.}
    \label{fig:Train-AARMSD}
\end{wrapfigure}

Predicting biomolecular complex structures is a central problem in drug discovery. 
In virtual screening, millions of candidate compounds must be rapidly evaluated by estimating their three-dimensional binding poses with respect to a target protein. 
This requirement for efficient and accurate complex structure prediction gives rise to the task of \textit{flexible docking}, where the goal is to predict both the ligand binding pose and the protein bound (holo) structure given the unbound (apo) protein structure and a ligand SMILES string. Current generative-based flexible docking methods \citep{diffdockpocket, redock, harmonicflow, flexdock, zhou2025energy} achieve state-of-the-art performance by iteratively denoising from random configurations to the target structure with Riemannian diffusion or flow matching \citep{riemanniandiff, riemannianfm}.

With recent advances in computational protein structure prediction, apo structures can be readily predicted using protein folding models such as AlphaFold~\citep{alphafold2} and ESMFold~\citep{esmfold}. Although the folding model provides an informative structure prior that simplifies the docking task by reducing it to modeling only the protein conformational shift, this dependence also introduces a potential source of data bias. Specifically, a mispredicted protein conformation can deviate significantly from its bound conformation. Such examples are effectively \textit{misspecified}, motivating either robust filtering methods \citep{flexdock} or substantially downweighting the long tail during training \citep{zhou2026robustdock}.

On the other hand, in practice, protein movement upon ligand binding can vary significantly. While most proteins only exhibit small flexibility within the pocket, large movements do occur and are often functionally meaningful, for instance, the formation of a cryptic pocket. This provides the second interpretation of the large conformational shift in flexible docking and motivates hard example mining, where the long tail structure should be upweighted during training to improve the generalization of the minority examples.

Figure~\ref{fig:Train-AARMSD} visualizes this effect that the all atom root mean square deviation (RMSD) between the apo and holo states in the PDBBind training data exhibits a long-tailed distribution, with most proteins undergoing only small conformational shifts and a small fraction exhibiting large structure changes. 
Practical evidence \citep{flexdock, zhou2026robustdock} shows that explicitly downweighting the large shift can improve the ligand docking performance but still remain limited in modeling large conformational shift. 
In this work, we seek for a solution to improve flexible docking performance without explicitly downweighting the long tail. To achieve this, we investigate using a curriculum training strategy for generative flexible docking that gradually incorporates large conformational examples. This implicitly downweights the long tail of challenging large shift cases during optimization, while still exposing the model to rare but genuine instances of substantial conformational change. Empirically, we find that a simple linear pacing function is most effective, lifting the top-10 ligand RMSD $<2$\,\AA{} success rate from 52.66\% to 56.14\% over a matched non-curriculum baseline. With relaxation and confidence-based selection, the resulting model GCLDock reaches 46.52\% top-1 success on the PDBBind ESMFold benchmark.

\section{Related Work}
\paragraph{Protein Ligand Flexible Docking.}Traditional docking searches an explicitly parameterized conformational space, and methods differ mainly in how receptor flexibility is encoded: discrete side-chain rotamers~\citep{leach1994}, selected residues appended to the search vector~\citep{trott2010autodockvina}, or side-chain rotamer search with backbone
relaxation~\citep{davis2009rosettaligand}. These remain competitive baselines, but their search cost grows sharply with the number of receptor degrees of freedom, which in practice restricts flexibility to a handful of pre-selected residues.
Generative models replace this search with learned sampling, specifically, DiffDock~\citep{diffdock} denoises the ligand pose from a random translation, rotation and set of torsion angles on the product manifold $\mathbb{R}^{3} \times SO(3) \times \mathbb{T}^{m}$. Subsequent work extends the same treatment to the receptor, first to side-chain torsions~\citep{redock, diffdockpocket} and then to the backbone via Riemannian flow matching~\citep{flexdock}. Following \citep{flexdock}, in this work, we use a diffusion process for the ligand pose generation and Riemannian flow matching for predicting the receptor conformational change.

\paragraph{Learning from Long-Tailed Distributions.} Long-tailed learning typically addresses class imbalance, where the training labels follow a skewed categorical distribution. Classical remedies include resampling~\citep{chawla2002smote}, reweighting~\citep{lin2017focal}, and transfer learning~\citep{yang2020rethinking}. The first two rebalance each class's
contribution during training, while the last imports additional information to offset the scarcity of tail examples. Flexible docking differs in two respects. First, the imbalanced quantity is continuous, so class-level resampling and reweighting do not apply directly. Second, and more importantly, these methods presuppose a supervised objective on the target itself, whereas our training signal is a conditional flow matching loss whose regression target is a velocity at a sampled time.

\section{Methods}
\label{sec:method}
\subsection{Curriculum Learning for Flexible Docking}
Curriculum learning (CL) \citep{bengio2009curriculum} is a training strategy that is inspired by the human learning process, where the training data is presented in order from easy to difficult. The model is first trained only on a fraction of easy data and gradually includes all the data points. It has been shown to improve model performance and accelerate training on a broad range of tasks, ranging from computer vision to natural language processing, etc. One of the motivations for CL is to "\textit{denoise}" where the difficult data is considered to be noisy or misspecified, and thus its influence on training should be reduced and only visible to the model until the late training stage \cite{wang2021surveyCL}. The model primarily learns the representation from the easy examples, which reflects the general patterns of the true data distribution. 
From this perspective, CL aligns with the practical evidence that reducing the training impact of large conformational examples can improve generalization. A key advantage of CL is that it does not discard large shift events, which may contain both challenging and noisy data. Besides, because only a subset of training examples is used during the early curriculum stages, the curriculum strategy can also reduce training computation.
Practically, CL requires devising two factors: a difficulty measurer and training scheduler. The former determines the definition of 'difficulty' for the training data, and the latter determines the speed of incorporating hard data into training. 

\paragraph{Difficulty Measurer.} As flexible docking predicts the conformational shift from apo to holo, we can directly adopt the shift RMSD as the difficulty measurer. This highlights the unique advantage of applying CL in this task where a natural difficulty measurer is off-the-shelf. This falls into the \textit{Predefined CL}, where the difficulty measurer is not learned by the neural network \citep{wang2021surveyCL}. 
We choose to use the global \textbf{C$\alpha$-RMSD}. This quantity remains invariant to the noise level and can capture the difficulty of protein motions outside of the pocket, such as hinge motion. Compared with full protein all-atom (AA) RMSD, C$\alpha$-RMSD more directly reflects the magnitude and difficulty of backbone conformational changes, which can be obscured in AA RMSD by the much larger number of side chain atoms.

\paragraph{Scheduler.} For the training scheduler, we aim to compare several pacing functions governing the speed of incorporation, encompassing the linear, geometric \citep{geoprogressionfunc} and root-p functions \citep{CLroot}. We limit ourselves to the continuous scheduler to avoid too many additional hyperparameters of discrete schedulers. We explain these three functions as follows:
\begin{itemize}
    \item geometric progression function: $\lambda_{\text{geom}}(t)=\min\left(1,2^{\left(\frac{\log_2 1-\log_2 \lambda_0}{T_{\text{grow}}}\cdot t+\log_2 \lambda_0\right)}\right).$
    \item linear function: $\lambda_{\text{linear}}(t)=\min\left(1,\frac{1-\lambda_0}{T_{\text{grow}}}\cdot t+\lambda_0\right).$
    \item root-p function: $\lambda_{\text{root-p}}(t)=\min\left(1,\sqrt{\frac{1-\lambda_0^p}{T_{\text{grow}}}\cdot t+\lambda_0^p}\right).$
\end{itemize}

where $t$ denotes the current training epoch, $T_{\mathrm{grow}}$ is the epoch number at which all data points are used for training for the first time, $\lambda_0$ is the initial fraction, and $p$ controls the curvature of the root-$p$ schedule. $\lambda(t)$ is the fraction of training examples at epoch $t$, increasing from $\lambda_0$ to $1$ according to the corresponding pacing function and remains at $1$ thereafter. Among these three, the linear function remains to be most intuitive, as the data points are gradually incorporated at a constant speed. The root-p function is proposed to give more time for training the newly added examples, with p for tuning the speed. Conversely, the geometric progression function gives more time for training on easy examples.

\subsection{Relaxation and Confidence Model}
While curriculum learning is only integrated for training the docking model, the docking produced poses usually do not satisfy the common energy constraints. Hence, we train a relaxation model with the empirical flat-bottom potential for energy minimization following \citep{flexdock}. This flat-bottom potential softly penalizes deviations from the target structure beyond a predefined tolerance. Moreover, the generative docking model produces multiple samples for each protein–ligand complex, from which the final prediction should be selected. For this, we also train the confidence model on the docking model inferred poses on the training set with the pairwise ranking loss \citep{ranknet, zhou2026robustdock}, with ranking labels derived from comparing the RMSD between the docking output and ground truth ligand structure. At inference, we sample poses with the docking model, refine them with the relaxation model, and select the pose with the highest ranking score.

\section{Experiments}
\paragraph{Data.} The model is trained on the well-known PDBBind 2020 \citep{pdb} dataset of timesplit, which consists of 16372 training data points, 968 validation and 363 test examples. The apo structure is generated by ESMFold with the sequence input. The dataset is preprocessed following \citep{diffdockpocket, flexdock}, where conformer matching is performed with the crystal structure to avoid the test-time distribution shift for the RDKit ligand conformer \citep{torsionaldiff}. The resulting conformer matched poses are used as the ground truth training signals. The receptor graph node attributes are initialized with the residue type and ESMFold embedding. 

\paragraph{Baselines.} We first investigate the effectiveness of curriculum learning (CL) against a non-CL baseline. Following \citep{diffdock, flexdock}, we build a self-constructed baseline, denoted as \textbf{FM Baseline}, which combines a DiffDock-style variance-exploding diffusion model for ligand generation with a Riemannian flow matching model for predicting protein conformational shift. We then incorporate CL into this baseline to assess its impact on docking accuracy. Using the optimal CL scheduler, we further train the relaxation and confidence models and evaluate the resulting method on the full PDBBind ESMFold Apo docking benchmark. On this, we compare against both traditional docking methods, SMINA \citep{koes2013SMINA} and GNINA \citep{mcnutt2021gnina}, and deep learning-based methods, including DiffDock-Pocket \citep{diffdockpocket}, Redock \citep{redock}, and FlexDock \citep{flexdock}.

\paragraph{Learning Rate Decay and Checkpoint Scheme.} For the non-CL baseline, we follow the previous practice \citep{zhou2026robustdock} where the learning rate is decayed based on the validation inference performance of top-5 ligand RMSD $<$ 2 \AA. For each validation example, this scheme loads the EMA checkpoint and performs 10 steps of denoising of five samples. The learning rate is decayed by a factor of 0.7 if there is no improvement in this metric for 30 epochs. The validation inference frequency is set to 10 epochs since it's a relatively costly computational process. However, with the CL, this learning rate scheduler is no longer valid. Because the validation metrics would change due to either newly added examples or genuine performance saturation. To address this, we first train our baseline and observe that the training converges with roughly 750 epochs, and then adopt the cosine annealing learning rate scheduler for learning rate decay of CL with a total of 800 epochs. In both cases, the last checkpoint is taken for evaluating the test performance.

\begin{figure*}[t]
    \centering
    \includegraphics[width=\textwidth]{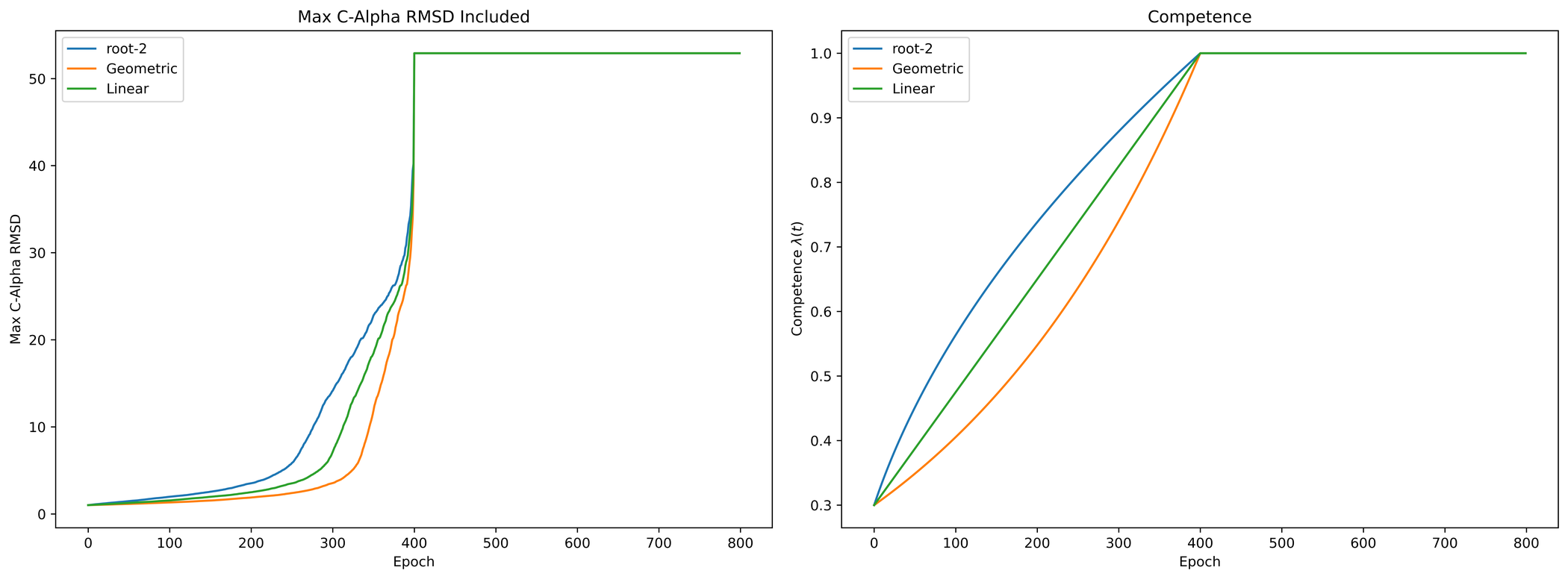}
    \caption{Max C-alpha RMSD and competence $\lambda(t)$ of curriculum learning at each epoch.}
    \label{fig:rmsd_competence}
    \vspace{-3.8mm}
\end{figure*}

\paragraph{Configurations of CL.} With a total of 800 training epochs, we set $T_{\mathrm{grow}}$ to be 400 epochs so that the large conformational examples are given at least half of the training time. The initial fraction $\lambda_0$ is set to 0.3, which is practically motivated to cover all the protein-ligand complexes with apo-holo shift less than 1 \AA. The competency and max C$\alpha$-RMSD included at each epoch of different schedulers are shown in Fig. \ref{fig:rmsd_competence}.

\begin{table}[t]
\centering
\caption{Top-10 and average PDBBind ESMFold docking benchmark. Results are averaged over 5 sampling seeds.}
\label{tab:rmsd_success}
\setlength{\tabcolsep}{7pt}
\renewcommand{\arraystretch}{1.15}

\begin{tabular}{lcccc}
\toprule
Method &
\multicolumn{2}{c}{Ligand RMSD $< 2$ \AA} &
\multicolumn{2}{c}{AA RMSD $< 1$ \AA} \\
\cmidrule(lr){2-3}
\cmidrule(lr){4-5}
& Top-10 & Average & Top-10 & Average \\
\midrule

FM Baseline
& 52.66 $\pm$ 0.32
& 37.57 $\pm$ 0.60
& \textbf{47.80 $\pm$ 0.71}
& 41.58 $\pm$ 0.51 \\
\midrule
\quad Geometric
& 51.82 $\pm$ 0.88
& 36.98 $\pm$ 0.29
& 46.41 $\pm$ 0.52
& 41.68 $\pm$ 0.42 \\

\quad Linear
& \textbf{56.14$\pm$ 0.84}
& \textbf{40.78 $\pm$ 0.43}
& 47.00 $\pm$ 0.26
& \textbf{42.12 $\pm$ 0.25} \\

\quad Root ($p=2$)
& 54.60 $\pm$ 0.69
& 37.13 $\pm$ 0.82
& 47.57 $\pm$ 0.50
& 41.44 $\pm$ 0.51 \\

\bottomrule
\end{tabular}
\label{tab:scheduler_compare}
\vspace{-2mm}
\end{table}

\paragraph{Evaluation Metrics.} To evaluate the effectiveness of CL, we first compare with the non-CL baseline on the common docking metrics: ligand RMSD and all-atom RMSD. We report both the fraction of top-10 (oracle) and average RMSD less than $c$ \AA. Following \citep{flexdock, redock}, $c$ is chosen to be 2 and 1 for ligand and protein, respectively. Once the optimal scheduler is determined, we further train the relaxation and confidence model. For the ultimate PDBBind performance, we further report the PB-Valid and ligand RMSD $<2$ ~\AA ~\& PB-Valid. The PB-Valid evaluates whether the produced structure pass the PoseBusters check \citep{posebusters}. Particularly, whether the intramolecular potential of the ligand and the intermolecular potential of the contact are satisfied, for example, the van der Waals force among the contact atoms and covalent bond length of ligand (Details see \citep{posebusters}).

\subsection{Results}
\paragraph{Results without Relaxation.} We first compare the docking accuracy achieved by different curriculum learning schedulers against our self-constructed baseline. Table \ref{tab:scheduler_compare} shows that with the linear scheduler, the docking accuracy improves by $+3.48 \%$ and $+3.21 \%$ for top-10 and average ligand RMSD $<$ 2 \AA, respectively, with protein prediction accuracy being largely retained. The results also show that \emph{incorporation speed is important}. The geometric progression function degrades predictive accuracy, while the root-2 function yields a smaller gain than the linear function. Notably, this improvement is achieved with reduced training compute (See Appendix \ref{app:compute}).

\paragraph{Results on PDBBind.} To further evaluate the effectiveness of curriculum learning for flexible docking, we train the relaxation and confidence models on top of the docking generative model using the linear scheduler, yielding \textbf{GCLDock}. As shown in Table~\ref{tab:top1-pdbbindbenchmark}, GCLDock substantially outperforms the baselines, achieving a $+6.82\%$ percentage points over FlexDock in the fraction of poses with ligand RMSD $<2$~\AA. Furthermore, the fraction of docking poses that simultaneously pass the PoseBusters check and achieve ligand RMSD $<2$~\AA{} increases from $33.7\%$ for the baseline to $40.26\%$, corresponding to a $+6.56$ percentage point improvement. Meanwhile, protein prediction accuracy is largely preserved relative to the baseline. These results demonstrate the effectiveness of curriculum learning in improving flexible docking performance.

\begin{table*}[t]
\caption{Top-1 PDBBind ESMFold docking benchmark. Results are averaged over 5 sampling seeds.}
\label{tab:top1-pdbbindbenchmark}
\vspace{-0.6em}
\centering
\small
\resizebox{\textwidth}{!}{%
\begin{tabular}{lccccc c}
\toprule
Method
& \multicolumn{2}{c}{Ligand RMSD}
& All-Atom RMSD
& \% PB valid
& \begin{tabular}[c]{@{}c@{}}L-RMSD $< 2$\,\AA{} \\ and \% PB Valid\end{tabular}
& Runtime (s) \\
\cmidrule(lr){2-3}
& \% $< 2$\,\AA{} ($\uparrow$)
& Median (\AA{}) ($\downarrow$)
& \% $< 1$\,\AA{} ($\uparrow$)
& ($\uparrow$)
& ($\uparrow$)
& \\
\midrule
SMINA (rigid)           & 6.6  & 7.7 & N.A. & --   & --  & 258  \\
SMINA                   & 3.6  & 7.3 & 5.2  & --   & --  & 1914 \\
GNINA (rigid)           & 6.7  & 7.1 & N.A. & \textbf{93.3} & 6.7 & 260 \\
GNINA                   & 8.4  & 7.9 & 4.5  & 91.3 & 7.7 & 1575 \\
\midrule
DiffDock-Pocket (rigid) & 37.5 & 3.0 & N.A. & 25.9 & 4.9 & 17 \\
DiffDock-Pocket         & 41.8 & 2.5 & 32.4 & 30.1 & 5.9 & 17 \\
ReDock                  & 39.0 & 2.5 & 39.8 & --   & --  & 15 \\
FlexDock                & 39.7 & 2.5 & \textbf{41.7} & 72.9 & 33.7 & 11 \\
\midrule
GCLDock
& $\mathbf{46.52 \pm 0.83}$
& $\mathbf{2.2 \pm 0.05}$
& 41.10 $\pm$ 1.02
& 79.89 $\pm$ 0.86
& $\mathbf{40.26 \pm 1.05}$
& $\mathbf{10.3}$ \\
\bottomrule
\end{tabular}
}
\end{table*}

\section{Conclusion}
In this work, we propose a curriculum learning strategy for generative-based flexible docking, motivated by two observations: 1) existing approaches improve docking performance by downweighting the training impact of large conformational shifts \citep{flexdock, zhou2026robustdock}, and 2) discarding such examples may remove genuine cases involving large conformational changes. With a carefully designed curriculum scheduler, our approach achieves improved docking performance with reduced training compute, reaching a top-1 ligand RMSD $<2$~\AA{} of $46.52\%$ on the PDBBind ESMFold Docking Benchmark. A limitation of our approach is that curriculum learning still does not explicitly distinguish genuine large conformational shifts from noisy predictions of ESMFold. More broadly, we believe curriculum learning can be readily extended to other protein modeling tasks in which example difficulty varies naturally with protein size or other structure complexity metrics.

\begin{ack}
This work was partially supported by the Knut and Alice Wallenberg Foundation. This work was in part financially supported by Digital Futures. 
The computations/data handling were enabled by resources provided by the National Academic Infrastructure for Supercomputing in Sweden (NAISS) at the Arrhenius supercomputer, hosted by Linköping University.
Experiments were performed using the supercomputing resource Berzelius provided by the National Supercomputer Centre at Linköping University and the Knut and Alice Wallenberg foundation.
\end{ack}


\bibliographystyle{plain}  
\bibliography{references}  


\appendix

\newpage

\section{Additional Results}

\paragraph{Stratified Results.} To further evaluate the gain of CL compared to the baseline, we stratify the PDBBind testset based on the pocket-aligned RMSD. This is computed by aligning the pocket residues of apo structure to the holo structure. The pocket residues are defined as the ones that have the minimum distance to any ligand atoms less than 5 \AA. As shown in Table \ref{tab:strategy_pdbbindtest}, CL training with a linear pacing function consistently improves all metrics, regardless of conformational flexibility. Interestingly, for the large conformational shift examples, the recovery of the ligand binding pose is significantly better than the protein-bound structure (27.0 $\%$ versus 7.3 $\%$).

\begin{table}[t]
  \caption{PDBBind test results stratified by the pocket-aligned apo-holo all-atom RMSD. The results are averaged over five sampling seeds.}
  \label{tab:strategy_pdbbindtest}
  \centering
  \small
  \setlength{\tabcolsep}{4pt}
  \begin{tabular}{l ccc ccc}
    \toprule
    & \multicolumn{3}{c}{Pocket-Alignd Apo-Holo Shift $<2$ \AA}
    & \multicolumn{3}{c}{Pocket-Alignd Apo-Holo Shift  $\geq2$ \AA } \\
    \cmidrule(lr){2-4}\cmidrule(lr){5-7}
    Method
    & \makecell{Top-10\\ligand \ $<2$ \AA} & \makecell{Avg.\\ligand \ $<2$ \AA} & \makecell{Avg. AA \\ RMSD $<1$ \AA}
    & \makecell{Top-10\\ligand \ $<2$ \AA} & \makecell{Avg.\\ligand\ $<2$ \AA} & \makecell{Avg. AA \\ RMSD $<1$ \AA} \\
    \midrule
    FM Baseline
      & 58.8 $\pm$ 1.2 & 43.7 $\pm$ 0.7 & 58.7 $\pm$ 0.2 
      & 40.1 $\pm$ 1.8 & 24.6 $\pm$ 0.6 & 6.2 $\pm$ 0.3 \\
    Geometric
      & 56.2 $\pm$ 1.7 & 42.6 $\pm$ 0.4 & 58.4 $\pm$ 0.4
      & 42.8 $\pm$ 1.7 & 25.3 $\pm$ 0.7 & 7.4 $\pm$ 0.3 \\
    Linear
      & \textbf{62.1} $\pm$ 1.2 & \textbf{47.7} $\pm$ 0.5 & \textbf{59.0} $\pm$ 0.2
      & \textbf{44.2} $\pm$ 0.7 & \textbf{27.0} $\pm$ 1.2 & \textbf{7.6} $\pm$ 0.3 \\
    Root ($p=2$)
      & 60.2 $\pm$ 0.3 & 44.0 $\pm$ 0.8 & 58.2 $\pm$ 0.5
      & 43.3 $\pm$ 2.3 & 23.2 $\pm$ 1.1 & 6.9 $\pm$ 0.3 \\
    \bottomrule
  \end{tabular}
\end{table}

\section{Training Compute Accounting}
\label{app:compute}

Curriculum learning changes the amount of data seen per epoch, so epoch counts are not
a meaningful basis for comparison. A competence-based curriculum trains on a growing
subset of the training set and therefore performs strictly less work per epoch than a
non-curriculum baseline until the curriculum saturates. We therefore report
\emph{complexes processed}, the total number of protein--ligand complexes for which a
forward and backward pass was computed over the course of training.

All runs in this comparison share an identical architecture ($71{,}966{,}718$ parameters; $n_s=50$, $n_v=10$, six convolution layers), an identical batch size ($12$ per device across $4$ devices) and an identical training set, so training compute is exactly proportional to complexes processed and the ratios in Table~\ref{tab:compute} are exact. We obtain the count directly from each run's checkpoint as $\texttt{global\_step} \times 12 \times 4$, and verify it against the analytic curriculum sum $\sum_{e} \max(1, \lfloor N c(e) \rfloor)$; predicted and recorded step counts agree to within a factor of $1.002$ for every run.

\begin{table}[t]
\caption{Training compute and docking accuracy for the pacing comparison. Accuracy is on the $363$ complex PDBBind timesplit test set, averaged over five sampling seeds. Every curriculum run consumes less compute than the baseline despite training for more epochs.}
\label{tab:compute}
\centering
\small
\begin{tabular}{lrrrrr}
\toprule
& \textbf{Epochs} & \textbf{Complexes} & \textbf{Rel.} &
\textbf{RMSD $<2$\,\AA{}} & \textbf{Median RMSD} \\
 & & \textbf{processed} & \textbf{compute} & \textbf{(\%)} & \textbf{(\AA{})} \\
\midrule
FM baseline      & 750 & 12{,}240{,}000 & $1.000\times$ & 52.7 & 1.87 \\
\midrule
Geometric            & 800 & 10{,}322{,}160 & $0.843\times$ & 51.8 & 1.92 \\
Root ($p=2$)         & 800 & 11{,}179{,}152 & $0.913\times$ & 54.6 & 1.81 \\
Linear               & 800 & 10{,}769{,}712 & $0.880\times$ & \textbf{56.1} & \textbf{1.77} \\
\bottomrule
\end{tabular}
\end{table}

\paragraph{Result.}
The linear scheduler Pareto-dominates the baseline, reaching $56.1\%$ against $52.7\%$ using $0.88\times$ the training compute; the root schedule likewise improves accuracy at $0.91\times$. The comparison is conservative, since matching the baseline to $800$ epochs would lower the linear schedule's relative cost to $0.825\times$. 

\paragraph{Absolute cost.}
For reference, one forward and backward pass over a single complex costs approximately
$663$\,GFLOP for this architecture, placing the baseline run at roughly $8.1$\,EFLOP of
training compute. This figure is a \emph{lower bound}. It is obtained with PyTorch's
\texttt{FlopCounterMode}, which accounts for \texttt{aten} matrix and convolution
operations but not for the fused cuEquivariance tensor-product kernels used by the
equivariant convolutions. It also varies by roughly $\pm 7\%$ with the size of the
sampled complexes. The relative column of Table~\ref{tab:compute} is unaffected by
either limitation, since the uncounted kernels scale identically across runs.

%
%
%

\section{Training Configuration}
In this section, we disclose our training details of docking model:
\begin{table}[ht]
\caption{Training configuration of the curriculum (linear pacing) docking model. The score network is an SE(3)-equivariant tensor-product message passing network operating on a ligand graph, protein residues graph, protein all-atom graph, protein-ligand graph with flexible backbone and side chains. Ligand translations, rotations and torsions each carry an independent noise schedule; protein backbone and sidechain degrees of freedom use fixed noise levels.}
\label{tab:training-config-linear}
\centering
\small
\begin{tabular}{lll}
\toprule
\textbf{Group} & \textbf{Hyperparameter} & \textbf{Value} \\
\midrule
\multirow{9}{*}{Architecture}
  & Total parameters                        & $72$M \\
  & Scalar / vector channels ($n_s$, $n_v$) & $50$ / $10$ \\
  & Convolution layers                      & $6$ \\
  & Spherical harmonics degree              & $2$ \\
  & Max irrep order                         & $2$ (second-order reps) \\
  & Sigma / distance / cross-distance embed & $64$ / $64$ / $64$ \\
  & Normalization                           & LayerNorm \\
  & Dropout                                 & $0.1$ \\
  & Tensor-product backend                  & cuEquivariance (fused) \\
\midrule
\multirow{7}{*}{Graph construction}
  & Ligand radius / max neighbours          & $5.0$\,\AA{} / --- \\
  & Receptor ($C_\alpha$) radius / neighbours & $15$\,\AA{} / $24$ \\
  & Protein atom radius / neighbours        & $5$\,\AA{} / $12$ \\
  & Protein-ligand cross graph cutoff         & $30$\,\AA{} \\
  & Ligand bond edge features               & $5$ \\
  & Protein language model                  & ESM-2 3B \\
  & Representation                          & all-atom, flexible bb.\ + sc. \\
\midrule
\multirow{7}{*}{Noise schedule}
  & Translation $\sigma$ range              & $[0.1, 5.0]$\,\AA{} \\
  & Rotation $\sigma$ range                 & $[0.1, 1.65]$\,rad \\
  & Torsion $\sigma$ range                  & $[0.0314, 3.14]$\,rad \\
  & Side chain torsion $\sigma$              & $0.1$\,rad \\
  & Backbone translation / rotation $\sigma$ & $0.1$\,\AA{} / $0.0$\,rad \\
  & Time sampling                           & $t \sim \mathcal{U}(0,1)$ \\
\midrule
\multirow{3}{*}{Curriculum}
  & Pacing function                         & linear \\
  & Initial competence $\lambda_0$                & $0.3$ \\
  & Curriculum length $T_{\text{grow}}$                   & $400$ epochs \\
\midrule
\multirow{9}{*}{Optimization}
  & Optimizer                               & AdamW \\
  & Learning rate                           & $1\times10^{-3}$ \\
  & Weight decay                            & $0.06$ \\
  & LR schedule                             & cosine annealing \\
  & \quad $T_{\max}$ / $\eta_{\min}$        & $800$ / $1\times10^{-5}$ \\
  & Batch size (per GPU $\times$ GPUs)      & $12 \times 4 = 48$ \\
  & Gradient accumulation                   & $1$ \\
  & EMA decay                               & $0.999$ \\
  & Precision                               & bfloat16 mixed \\
\midrule
\multirow{4}{*}{Schedule}
  & Training epochs                         & $800$ \\
  & Validation inference every              & $10$ epochs \\
  & Inference steps (sampling)              & $20$ \\
\bottomrule
\end{tabular}
\end{table}

\end{document}